\documentclass{article}

\usepackage[main,final]{neurips_2026}

\usepackage[utf8]{inputenc}
\usepackage[T1]{fontenc}
\usepackage{hyperref}
\usepackage{url}
\usepackage{booktabs}
\usepackage{amsfonts}
\usepackage{amsmath}
\usepackage{amssymb}
\usepackage{amsthm}
\usepackage{nicefrac}
\usepackage{microtype}
\usepackage{xcolor}
\usepackage{graphicx}
\usepackage{subcaption}
\usepackage{algorithm}
\usepackage{algorithmic}
\usepackage{multirow}

\newtheorem{theorem}{Theorem}
\newtheorem{assumption}{Assumption}
\newtheorem{proposition}{Proposition}

\newcommand{\eg}{\textit{e.g.}}

\title{RecMem: Recurrent Memory Compression for Long-Sequence Generative Recommendation}

\hypersetup{
  pdftitle={RecMem: Recurrent Memory Compression for Long-Sequence Generative Recommendation},
  pdfauthor={Haoyi Hu, Gaoyang Guo, Yunjia Xi, Wen Chen, Weiwen Liu, Jian Wu, Yuning Jiang, Weinan Zhang}
}

\author{%
  \textbf{Haoyi Hu}$^{1}$\thanks{Equal contribution.}\quad
  \textbf{Gaoyang Guo}$^{2}$\footnotemark[1]\quad
  \textbf{Yunjia Xi}$^{1}$\quad
  \textbf{Wen Chen}$^{2}$\thanks{Corresponding authors: Wen Chen and Weiwen Liu.}\\
  \textbf{Weiwen Liu}$^{1}$\footnotemark[2]\quad
  \textbf{Jian Wu}$^{2}$\quad
  \textbf{Yuning Jiang}$^{2}$\quad
  \textbf{Weinan Zhang}$^{1}$\\
  \normalfont $^{1}$Shanghai Jiao Tong University\quad $^{2}$Taobao \& Tmall Group of Alibaba\\
  {\small\texttt{hu\_haoyi@sjtu.edu.cn}\quad
  \texttt{guogaoyang.ggy@taobao.com}}\\
  {\small\texttt{chenyu.cw@taobao.com}\quad\texttt{wwliu@sjtu.edu.cn}}
}

\begin{document}

\maketitle

\begin{abstract}
Generative recommendation benefits from long interaction histories, but the cost
of processing them directly grows with history length. We propose
\textbf{RecMem}, a recurrent memory compression framework guided by three desiderata:
\emph{space invariance}, \emph{semantic fidelity}, and \emph{lifelong evolvability}.
RecMem compresses historical segments into a fixed set of memory slots using a
bifurcated encoder--decoder with shared lower layers and slot-wise forget-gated
updates. Recommendation supervision trains the memory to preserve predictive
information, while stream-consistency training aligns predictions from sequential
and packed processing. By construction, the memory state and decoder context are
independent of history length; under contraction and approximate semigroup
consistency, we further derive a horizon-independent bound on the state
discrepancy between these two processing orders. On MerRec, RecMem uses 16 memory
slots and a 216-token decoder context, compared with 1\,200 tokens for HSTU-1200,
while retaining 87.9\% of HSTU-1200's Recall@10 and 97.0\% of its Recall@200. It
outperforms PHSTU and the evaluated recent-only and retrieval baselines, and
analyses of distant history show that archived interactions contribute predictive
signal.
\end{abstract}

\section{Introduction}

Longer user behavior sequences expose richer and multi-faceted preferences invisible from
short interaction windows, and modern recommender systems (RSs) therefore benefit
from increasingly long histories~\citep{zhai2024actions,chai2025longer}. Recently, Transformer-based~\citep{vaswani2017attention} generative recommendation has reshaped the field by formulating personalization as a sequence-to-sequence generation task~\citep{lin2023can,zhai2024actions}. By treating historical interactions as structured contexts, generative recommenders can directly decode the user's next intent, fostering a more holistic understanding of user preference. 

However, this paradigm faces a formidable bottleneck: \textit{its computational and memory costs grow rapidly with sequence length}. Naively
feeding the entire user history, \eg, thousands of items in large-scale RSs~\citep{hou2024cross,xi2025bursting}, into a Transformer incurs context
explosion: the number of processed tokens grows with history length, GPU memory consumption rises accordingly, and inference latency becomes prohibitive. 
To incorporate long-term user history into generative recommendation with computational efficiency, it is imperative to move beyond raw sequence modeling towards \textbf{memory mechanisms}, which compress long user interaction
histories into compact preference memory slots~\citep{zhang2025efficient,chen2026recurrent}.

The central challenge is to compress growing interaction histories while
retaining information useful for recommendation. We argue that an ideal memory
mechanism should satisfy three properties simultaneously:
\begin{enumerate}
\item \textbf{Space Invariance.} The retained memory should occupy $O(1)$ space
with respect to history length. With a fixed recent window, prediction should
use a decoder context whose size is also independent of the number of processed
segments.
\item \textbf{Semantic Fidelity.} The memory should faithfully retain rich
information from the full user history, including long-term interests and
structural behavioral patterns predictive of future interactions. It should
support recommendation quality similar to that obtained from the full history.
\item \textbf{Lifelong Evolvability.} The memory should dynamically update its
internal state and remain stable and usable as the history grows without bound.
Recurrent compression introduces per-step approximation errors that could
accumulate over many updates; an ideal memory mechanism should keep these
errors controlled.
\end{enumerate}

Existing approaches face different obstacles to jointly satisfying these
desiderata. Fixed-window models select recent
interactions~\citep{kang2018self,sun2019bert4rec,zhai2024actions}, while retrieval
methods select relevant historical interactions~\citep{lin2024rella,qin2020user};
both operate on a subset of the available history. This selection limits
semantic fidelity when excluded interactions contain predictive clues about
user preferences. Segment-based methods such as
PersonalizedHSTU (PHSTU)~\citep{zhang2025efficient} summarize historical segments
into user-token/KV states whose size grows with the number of retained segment
summaries, which violates space invariance. Teacher-guided methods
such as Rec2PM~\citep{chen2026recurrent} learn
memory compression using teacher-generated reference states. In contrast, we aim
to combine a fixed recurrent state with direct recommendation supervision and an
explicit analysis of processing-order stability, without relying on a teacher
model.

To address these requirements, we present \emph{RecMem}, a recurrent memory
compression framework for generative recommendation. \textit{First}, each
historical segment updates a fixed set of $M$ memory slots through a slot-wise
forget-gated rule. The decoder reads these slots together with a fixed recent
window, so both the retained state and the decoder context are independent of
history length (space invariance).
\textit{Second}, a bifurcated encoder--decoder shares lower layers and uses
separate upper blocks for memory encoding and next-item decoding. Recommendation
supervision trains the resulting fixed-capacity bottleneck to retain predictive
historical signals, targeting semantic fidelity. \textit{Finally},
stream-consistency training aligns the recommendation distributions produced by
sequential and packed processing. Under explicit assumptions of contraction and
approximate semigroup consistency, our analysis bounds the state discrepancy
between these processing orders independently of history length, targeting
lifelong evolvability. We evaluate predictive retention using
recommendation experiments and distant-history controls, and examine update
behavior through empirical diagnostics.

Our contributions are as follows:
\begin{itemize}
\item We identify three essential desiderata for ideal memory mechanisms in
generative recommendation: \textbf{space invariance}, \textbf{semantic fidelity},
and \textbf{lifelong evolvability}.
\item We propose \emph{RecMem}, a recurrent memory framework combining a
bifurcated encoder--decoder, slot-wise forget-gated updates, and stream-consistency
training, whose retained state and decoder context have a fixed size.
\item We establish a history-length-independent bound on the state discrepancy
between sequential and packed processing under contraction and approximate
semigroup consistency, and interpret the fixed-capacity memory as a bottleneck
trained for next-item prediction.
\item On MerRec, we quantify the accuracy--context trade-off, demonstrate
predictive gains from archived history, and use shared-stack and
near-parameter-matched controls to show that RecMem's gains are not explained
by additional parameters. Evaluating a single trained RecMem checkpoint on a fixed
user cohort, without retraining, shows stable aggregate recommendation quality
through 10\,000 interactions.
\end{itemize}

\section{Related Work}
\label{sec:related}


\paragraph{Generative Recommendation.} 
Large language models (LLMs) have substantially influenced recommendation research~\citep{lin2023can,xi2023towards}. Generative recommendation reformulates personalization as an end-to-end sequence-to-sequence generation task~\citep{geng2022recommendation,rajput2023recommender}. Representative backbones include text-to-text prompting frameworks~\citep{geng2022recommendation}, semantic-ID generative retrieval~\citep{rajput2023recommender}, and large Transformer architectures such as HSTU~\citep{zhai2024actions}, HLLM~\citep{chen2024hllm}, and other long-sequence LLM recommenders~\citep{chai2025longer}. In these models, user behavior is consumed as model context, decoder-side tokens, or persisted token/KV states. However, scaling to lifelong histories stresses the same resource bottlenecks inherent in LLMs: attention costs, serving latency, and per-user storage grow prohibitively with the amount of visible history. RecMem targets this generative setting by compressing the archived prefix into a fixed set of token-like memory slots, while keeping the recent suffix at item-level resolution for next-item prediction.

\paragraph{Recommendation with Lifelong User Behaviors.}
Beyond the generative setting, long user histories have been studied extensively via attention-based interest extraction~\citep{zhou2018deep,zhou2019deep}, target-aware retrieval~\citep{pi2020search}, and search-based aggregation~\citep{chang2023twin}. These methods handle thousands of interactions by filtering behaviors before ranking, bypassing the autoregressive Transformer bottleneck entirely, and are thus complementary to our scope. Within the generative paradigm, PHSTU~\citep{zhang2025efficient} augments interaction streams with personalized summary tokens; its user-token/KV state grows with the number of summarized segments up to a maximum history length, beyond which older interactions are discarded. Conversely, Rec2PM~\citep{chen2026recurrent} decouples memory size from history length but relies on full-context teacher-model guidance, inheriting the teacher's rigid window as a scalability ceiling. Distinct from both, RecMem isolates generative backbones from the context wall by learning a self-contained recurrent update rule whose retained state has a strictly history-independent size.

\paragraph{Context and Recurrent Memory Compression.}
Token pruning and KV-cache compression~\citep{xiao2023efficient,mu2023learning,
rae2019compressive} reduce attention cost within a single forward pass.
Infini-attention~\citep{munkhdalai2024leave} maintains compressive memory across
discarded KV states. Recurrent memory transformers and related LLM compression
methods~\citep{bulatov2022recurrent,chevalier2023adapting,hutchins2022block} carry compact
state across segments, while gated recurrent networks~\citep{graves2012long,cho2014learning}, state
space models~\citep{gu2021efficiently,gu2023mamba}, retention networks~\citep{sun2023retentive},
LRURec~\citep{yue2024linear}, and Titans~\citep{behrouz2024titans} provide broader blueprints for
fixed-capacity sequence processing. RecMem adapts this principle to
generative recommendation: it produces a bounded user memory \emph{before}
decoding, so the decoder context remains fixed even as the historical prefix
continues to grow.

\section{Method}
\label{sec:method}

\subsection{Problem Formulation}\label{sec:formulation}
Given a generative recommender system, let $X = [i_1, i_2, \ldots, i_T]$ denote the full chronological interaction history of a user, where $i_t$ is the $t$-th interacted item and $T$ is the total sequence length. To model lifelong user preferences efficiently, we partition the unbounded history $X$ into two complementary parts: an archived history prefix $X_{\mathrm{pre}}$ and a recent interaction suffix $X_{\mathrm{rec}}$.
Specifically, the recent suffix $X_{\mathrm{rec}} = [i_{T-L_r+1}, \ldots, i_T]$ consists of the most recent $L_r$ items. The remaining historical items form the prefix $X_{\mathrm{pre}} = [i_1, \ldots, i_{T-L_r}]$, which is right-aligned and chunked into $K$ non-overlapping contiguous segments $[S_1, S_2, \ldots, S_K]$, each with a fixed maximum length $L_s$ (i.e., $K = \lceil (T-L_r)/L_s \rceil$).

We formalize memory-augmented generative recommendation as two stages: compression and prediction. The \textbf{compression stage} processes the segment sequence of $X_{\mathrm{pre}}$  recurrently. At each step $t \in [1, K]$, a memory module $U$ updates the compressed user state:
\begin{equation}\label{eq:def_memo}
M_t = U(M_{t-1}, S_t),
\end{equation}
where $M_t$ is the memory state after the first $t$ segments, and $M_0$ is initialized to a zero matrix. The archived prefix therefore enters prediction only through the final state $M_K$. Next, in the \textbf{prediction stage}, the recommendation backbone $\mathcal{F}$ takes the final memory state $M_K$ and the raw item embeddings of the recent suffix $X_{\mathrm{rec}}$ as inputs to decode the user's next intent:

\begin{equation}
\widehat{y}_{T+1} = \mathcal{F}(M_K, X_{\text{rec}}).
\end{equation}

We jointly optimize the recurrent memory module $U$ and the recommendation backbone $\mathcal{F}$ to retain predictive information from archived interactions within a fixed memory budget. Our design targets three desiderata: \emph{space invariance}, via a history-independent state and decoder context; \emph{semantic fidelity}, via retaining predictive information from the full history so as to approach full-history recommendation quality; and \emph{lifelong evolvability}, via stable recurrent updates whose rollout error remains controlled as $K \to \infty$. We analyze processing-order stability under explicit assumptions and evaluate predictive fidelity empirically.

\subsection{RecMem Architecture}

\begin{figure}[t]
\centering
\includegraphics[width=\linewidth]{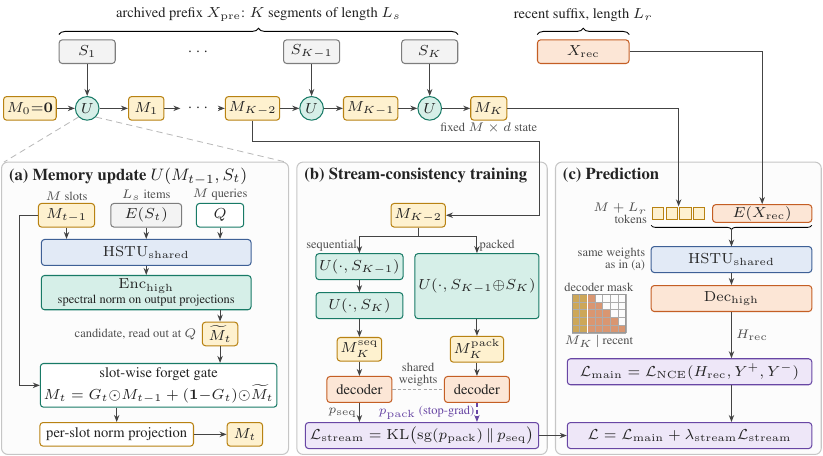}
\caption{Overview of RecMem. Top: $U$ folds the $K$ archived segments into a
fixed-size memory state, one segment at a time. (a)~In each update, the
slot-wise forget gate is computed from the mask-aware mean of $E(S_t)$.
(b)~Both branches start from the same $M_{K-2}$, and the packed update uses
repeated segment-local positions and a causal mask. The two final states are
decoded with the same $X_{\mathrm{rec}}$ and candidates; $\mathrm{sg}$ denotes
stop-gradient. (c)~Recent tokens attend to all memory slots and causally to
each other. The decoder outputs at all recent positions, $H_{\mathrm{rec}}$,
enter the NCE loss with shifted positives $Y^{+}$ and sampled negatives
$Y^{-}$; at inference, the last row of $H_{\mathrm{rec}}$ scores candidate
items. Blue, green, and orange mark shared, encoder-side, and decoder-side
components.}
\label{fig:framework}
\end{figure}

To operationalize the problem formulation, we present \emph{RecMem}, implemented as a recurrent extension of HSTU~\citep{zhai2024actions} using a \textbf{bifurcated encoder--decoder} architecture (Figure~\ref{fig:framework}). The backbone layers are decoupled into lower shared blocks ($\mathrm{HSTU}_{\mathrm{shared}}$) that capture general sequential dependencies, and specialized upper blocks ($\mathrm{Enc}_{\mathrm{high}}$, $\mathrm{Dec}_{\mathrm{high}}$) that separate memory encoding from next-item decoding.

The default configuration uses eight shared lower blocks, eight encoder-specific upper blocks, and eight decoder-specific upper blocks. It therefore contains 24 unique backbone blocks, with an active depth of 16 blocks on each of the encoding and decoding paths. We refer to this default configuration as SplitTop-8/8/8.

Memory compression (Figure~\ref{fig:framework}a) implements the recurrence in Eq.~\eqref{eq:def_memo}. For the $t$-th prefix segment $S_t$, we project its items into embeddings $E(S_t) \in \mathbb{R}^{L_s \times d}$ and introduce a learned bank of memory-query tokens $Q \in \mathbb{R}^{M \times d}$. The encoder constructs a concatenated input sequence $\mathbf{H}_t^{(0)} = [M_{t-1}; E(S_t); Q]$. Within $\mathrm{HSTU}_{\mathrm{shared}}$, a multi-head attention mechanism allows segment tokens $E(S_t)$ to attend to the previous memory $M_{t-1}$, while the query tokens $Q$ attend to both $M_{t-1}$ and the full segment $S_t$. After passing through the shared lower layers and the encoder-specific upper layers $\mathrm{Enc}_{\mathrm{high}}$, the intermediate representations at the designated $Q$ positions are extracted to form the candidate preference memory $\widetilde{M}_t$:
\begin{equation}\widetilde{M}_t = \mathrm{Enc}_{\text{high}}\Big(\mathrm{HSTU}_{\text{shared}}([M_{t-1}; E(S_t); Q])\Big).
\end{equation}


After all $K$ prefix segments have been compressed into $M_K$, the decoder (Figure~\ref{fig:framework}c) computes $\mathcal{F}(M_K, X_{\mathrm{rec}})$. The input sequence to the decoder is formed by concatenating the memory slots and recent item embeddings: $\mathbf{D}^{(0)} = [M_K; E(X_{\mathrm{rec}})]$. Within $\mathrm{HSTU}_{\mathrm{shared}}$ and $\mathrm{Dec}_{\mathrm{high}}$, a customized attention mask lets each recent token attend to all memory slots and, causally, to itself and the preceding recent tokens. The decoder output hidden states at the $L_r$ recent positions are:
\begin{equation}\label{eq:decoder_output}
H_{\mathrm{rec}} = \mathrm{Dec}_{\mathrm{high}}\!\Big(\mathrm{HSTU}_{\mathrm{shared}}(\mathbf{D}^{(0)})\Big)\Big|_{X_{\mathrm{rec}}},
\end{equation}
where $\big|_{X_{\mathrm{rec}}}$ extracts hidden states at the $L_r$ recent token positions. $H_{\mathrm{rec}} \in \mathbb{R}^{L_r \times d}$ is subsequently used to compute next-item generation probabilities. Because the archived prefix reaches the decoder only through $M_K$, the decoder context is fixed at $M + L_r$ tokens.

\subsection{Memory Update and Compression}

We update the memory using a slot-wise forget gate~\citep{cho2014learning}. Let $\bar{\mathbf{s}}_t \in \mathbb{R}^d$ denote the mask-aware mean of the segment item embeddings. The gate is computed as
\begin{equation}
\mathbf{g}_t = \sigma(W_g\bar{\mathbf{s}}_t + \mathbf{b}_g) \in (0,1)^M,
\qquad G_t = \mathbf{g}_t\mathbf{1}_d^{\top},
\end{equation}
where $W_g \in \mathbb{R}^{M \times d}$ and $\mathbf{b}_g \in \mathbb{R}^M$ are learnable parameters. Each slot receives its own scalar gate, which is broadcast across that slot's embedding dimensions. The memory update is
\begin{equation}
M_t = G_t \odot M_{t-1} + (\mathbf{1}_{M \times d} - G_t) \odot \widetilde{M}_t.
\end{equation}

If a segment $S_t$ contains only padding items, the gating mechanism bypasses the update and sets $M_t = M_{t-1}$. To avoid representation explosion during long-term recurrence, each memory slot is individually projected onto a Euclidean ball of radius $c$:
\begin{equation}
M_t^{(m)} \leftarrow \min\left(1, \frac{c}{\|M_t^{(m)}\|_2}\right) M_t^{(m)},
\end{equation}
where $m \in [1, M]$ indexes the memory slots, $M_t^{(m)} \in \mathbb{R}^d$ denotes the $m$-th row of $M_t$, and $c$ is the configured maximum norm bound.

In the default SplitTop-8/8/8 configuration, spectral normalization~\citep{miyato2018spectral} is applied to the output projection of each encoder-specific upper block. Position embeddings are local to each segment and to the recent suffix, so recurrent processing does not require an increasing absolute position index. These choices support the recurrent design; the contraction condition used in our stability analysis is stated explicitly in Section~\ref{sec:theory}.



\subsection{Training Objectives and Optimization}


Training begins from an all-zero initial state $M_0 = \mathbf{0}$. To expose training to realistic recurrent states, the first $K_{\mathrm{burn}}$ segments are processed without gradients. The resulting intermediate memory state is detached from the computational graph, after which the remaining $K_{\mathrm{roll}}$ segments participate in gradient-carrying rollout. RecMem incorporates a rollout curriculum: at predefined training steps, $K_{\mathrm{roll}}$ is progressively increased while keeping the total segment budget $K = K_{\mathrm{burn}} + K_{\mathrm{roll}}$ fixed, to stabilize gradients through long rollouts.

The primary recommendation objective is the NCE loss applied to the recent suffix $X_{\mathrm{rec}}$. For a training sequence, the recent window expands to include the true next-item label at the final position, denoted as $X_{\mathrm{rec}}^{+} = [i_{T-L_r+1}, \ldots, i_T, i_{T+1}]$. We define the shifted positive labels as $Y^{+} = [i_{T-L_r+2}, \ldots, i_{T+1}]$, which are aligned position-wise with the decoder outputs. For each position in $Y^{+}$, the training sampler draws a set of negative items from the item pool, denoted as $Y^{-}$.  Let $H_{\mathrm{rec}}$ be the decoder output defined in Eq.~\eqref{eq:decoder_output}. The main recommendation loss is defined as:
\begin{equation}
\mathcal{L}_{\mathrm{main}} =
\mathcal{L}_{\mathrm{NCE}}(H_{\mathrm{rec}}, Y^{+}, Y^{-}).
\end{equation}
Because the decoder cannot attend to the archived prefix directly, the archived history affects $\mathcal{L}_{\mathrm{main}}$ only through $M_K$; the resulting gradients train the encoder to retain task-relevant signals under the fixed-capacity bottleneck.



We introduce a stream-consistency objective (Figure~\ref{fig:framework}b) to align predictions obtained from sequential and packed processing of the same history. Starting from $M_{K-2}$, the sequential branch processes $S_{K-1}$ and $S_K$ in two recurrent updates, while the packed branch concatenates them into one update. The shared decoder maps each memory state, together with the same recent suffix, to next-item distributions $p_{\mathrm{seq}}$ and $p_{\mathrm{pack}}$ over an identical candidate set.

We choose the last two segments to place the consistency gradient close to the prediction step while keeping the packed computation small. Packing additional segments would increase the encoder input length and its memory cost.

The packed branch repeats the segment-local position indices as $[0,\ldots,L_s-1,0,\ldots,L_s-1]$. A causal mask over the combined block allows the second segment to attend to the first while preserving each item's within-segment position. The objective is
\begin{equation}
\mathcal{L}_{\mathrm{stream}} = \mathrm{KL}\!\left(
\operatorname{stopgrad}(p_{\mathrm{pack}}) \,\|\, p_{\mathrm{seq}}\right).
\end{equation}
Gradients from this loss propagate through the sequential computation. Two-segment packing serves as a local training proxy for processing-order consistency; the composition diagnostics in Appendix~\ref{app:stability} use the same construction.

The full joint optimization objective is formulated as:
\begin{equation}
\mathcal{L} = \mathcal{L}_{\mathrm{main}} + \lambda_{\mathrm{stream}} \mathcal{L}_{\mathrm{stream}},
\end{equation}
where $\lambda_{\mathrm{stream}}$ is a balancing hyperparameter. The complete training procedure is summarized in Appendix~\ref{app:algorithm}.

\subsection{Inference}
At inference time, the memory state $M$ is persistently cached across prediction steps. When a new interaction completes a segment, only the latest segment $S_{\mathrm{new}}$ is processed through a single update $M \leftarrow U(M, S_{\mathrm{new}})$, keeping the per-step compression cost constant regardless of history length. When no cached state is available (cold construction), the full prefix is partitioned into $N_{\mathrm{seg}}$ segments and compressed sequentially via $M_t = U(M_{t-1}, S_t)$ for $t = 1, \ldots, N_{\mathrm{seg}}$.

Given the current memory state $M$ and the most recent $L_r$ items $X_{\mathrm{rec}}$, the decoder conditions on $[M; E(X_{\mathrm{rec}})]$ and produces $H_{\mathrm{rec}}$ via Eq.~\eqref{eq:decoder_output}. The score for each candidate item $j$ is computed as:
\begin{equation}\label{eq:inference_score}
\mathrm{score}(j) = \mathbf{h}_T^\top \mathbf{e}_j,
\end{equation}
where $\mathbf{h}_T = H_{\mathrm{rec}}[L_r] \in \mathbb{R}^d$ is the last position's hidden state and $\mathbf{e}_j$ is the embedding of item $j$.


\section{Theoretical Analysis}
\label{sec:theory}

We analyze RecMem's memory properties at three levels. Fixed state size and
decoder context follow from the architecture. Under contraction and
approximate semigroup consistency, we derive a horizon-independent bound on
the state discrepancy between sequential and packed processing of the same
history. We then interpret the fixed-capacity memory as a predictive
bottleneck, whose retention of useful historical information is evaluated
empirically.

\subsection{Lifelong Evolvability: Conditional Rollout Stability}
\label{sec:rollout_stability}

At inference time, RecMem compresses user history by processing segments
sequentially. An alternative is to pack all segments into a single context and
process them in one pass. If the update operator were perfectly associative,
both approaches would yield identical memory states. In practice, the
segment-wise approximation introduces per-step errors that could accumulate
over segments. We show that these errors remain bounded under two explicit
assumptions on the update operator.

\begin{assumption}[Contraction]
\label{ass:contraction}
The update operator $U$ satisfies
$\sup_{M,S} \left\| \partial U(M,S) / \partial M \right\|_2 \le \rho$
for some $\rho \in [0,1)$.
\end{assumption}

The forget gate controls the contribution of the previous memory, while
spectral normalization regularizes selected encoder projections. The
contraction property concerns the complete update operator, including the
candidate memory's dependence on the previous state. We therefore treat
contraction as an explicit assumption and examine local update behavior
through empirical diagnostics.

\begin{assumption}[Approximate Semigroup Consistency]
\label{ass:semigroup}
For any memory state $M$, any packed prefix block $A$ formed by concatenating
consecutive segments, and the next contiguous segment $B$,
\[
\| U(U(M, A), B) - U(M, A \oplus B) \|_F \le \varepsilon,
\]
where $A \oplus B$ denotes the concatenation of $A$ and $B$, and
$\varepsilon \ge 0$ is the semigroup consistency gap.
\end{assumption}

Approximate semigroup consistency measures the state discrepancy caused by
regrouping the same history. The stream-consistency objective aligns the
resulting recommendation distributions. Because distinct memory states can
yield similar decoder outputs, reducing output KL alone does not establish
the state-space bound in this assumption. Moreover, training uses the last
two segments as a local proxy, whereas the assumption applies to arbitrary
packed prefixes and their next contiguous segment.

\begin{theorem}[Rollout Stability under Contraction and Approximate Semigroup Consistency]
\label{thm:rollout}
Under Assumptions~\ref{ass:contraction} and~\ref{ass:semigroup}, sequential
and packed processing of the same $K$ segments from the same zero initial
state satisfy
\[
\left\| M_K^{\mathrm{seq}} - M_K^{\mathrm{pack}} \right\|_F \le
\varepsilon\frac{1 - \rho^K}{1 - \rho} \le
\frac{\varepsilon}{1 - \rho}.
\]
\end{theorem}

\begin{proof}[Proof sketch]
Let $P_k = S_1 \oplus \cdots \oplus S_k$ denote the packed prefix
($P_0=\emptyset$) and define $M_k^{\mathrm{pack}} = U(\mathbf{0}, P_k)$, with
$M_0^{\mathrm{pack}} = \mathbf{0}$. Let
$e_k = \| M_k^{\mathrm{seq}} - M_k^{\mathrm{pack}} \|_F$ denote the rollout
error at step $k$. By the contraction property, the triangle inequality, and
Assumption~\ref{ass:semigroup} applied to the packed prefix $P_k$ and segment
$S_{k+1}$,
\[
e_{k+1} \le \rho \cdot e_k + \varepsilon.
\]
Unrolling from $e_0 = 0$ yields
$e_K \le \varepsilon \sum_{j=0}^{K-1} \rho^j \le \varepsilon / (1 - \rho)$.
The complete proof is provided in Appendix~\ref{app:proof_rollout}.
\end{proof}

Under the stated assumptions, contraction attenuates previously accumulated
discrepancy, while approximate semigroup consistency bounds the discrepancy
introduced by regrouping segments. Their combination yields a bound
independent of the number of updates. This result concerns processing-order
stability in memory space. Recommendation quality and retention of
distant-history information are evaluated separately.

\subsection{Semantic Fidelity: Fixed-Capacity Bottleneck}
\label{sec:ib}

The fixed-capacity memory in RecMem creates a practical bottleneck between the
archived history and the prediction target.
Let
$X_{\mathrm{pre}} = (S_1, \ldots, S_K)$ denote the segmented archived prefix. The memory
$M_K$ is a deterministic function of $X_{\mathrm{pre}}$ with fixed dimensionality
$M \times d$, where $M$ is the number of slots and $d$ is the embedding
dimension. Increasing the history length $K$ therefore does not increase the
number of memory coordinates or decoder-side memory tokens available to encode
old interactions. This property is particularly useful when the recommendation
backbone would otherwise expose long histories as decoder context, token
memory, or KV state.

\begin{proposition}[Fixed-Capacity Bottleneck]
\label{prop:ib}
For fixed $M$, $d$, and recent window length $L_r$, RecMem allocates
$M \times d$ recurrent state coordinates and $M + L_r$ decoder-side tokens
regardless of the number of processed segments $K$.
\end{proposition}

\begin{proof}
By construction, each segment update maps the previous state and the current
segment to $M$ slots of dimension $d$. The decoder always receives those
$M$ slots plus the $L_r$ recent item tokens. Neither quantity depends on
$K$.
\end{proof}

The proposition establishes the memory and decoder-context budgets. Within
this fixed capacity, recommendation supervision trains the recurrent state
to retain historical signals useful for predicting future interactions.
Since the decoder accesses the archived prefix only through $M_K$,
comparisons with a separately trained recent-only model and analyses of
archived-only recurrence targets directly assess the predictive value of
the stored history. We evaluate this empirical notion of semantic fidelity
on MerRec, alongside the memory-capacity ablations.

\section{Experiments}
\label{sec:experiments}

\subsection{Experimental Setup}

\paragraph{Dataset.}
We evaluate on MerRec~\citep{li2025merrec}, a large-scale benchmark collected from
the Mercari C2C marketplace\footnote{\url{https://huggingface.co/datasets/mercari-us/merrec}}. MerRec contains interactions from over 5 million users,
many with long histories, which makes it suitable for evaluating
long-context generative recommendation. We adopt a temporal split protocol following \citet{zhang2025efficient}:
each user's first 1\,200 interactions form the training set. At each test
position $t \ge 1200$, the model predicts the next item from the available
context up to position $t$.

\paragraph{Baselines.}
We compare HSTU-based recommenders that differ in how they represent interaction
history. \textbf{HSTU-1200}~\citep{zhai2024actions} applies full attention to the
most recent 1\,200 interactions, while \textbf{HSTU-FullPrefix} processes the
growing uncompressed prefix. Full-prefix attention preserves direct access to
history at a growing computational and memory cost, motivating compressed
alternatives for serving. \textbf{PHSTU}~\citep{zhang2025efficient} summarizes
historical segments into user-token/KV state. \textbf{HSTU-Recent-200} is a
separately trained control that receives only the most recent 200 interactions.
\textbf{Ret-HSTU-16} prepends 16 retrieved historical item occurrences to the
recent 200-item window, matching RecMem's 216-token decoder context. Retrieval
is restricted to the user's history before the recent window and has no access
to the target or future interactions (Appendix~\ref{app:retrieval}). Together,
these comparisons assess the trade-offs among explicit history, segment-level
compression, retrieval, and fixed recurrent memory. We report two RecMem
variants: \textbf{RecMem-Main}, trained with $\mathcal{L}_{\mathrm{main}}$ only,
and \textbf{RecMem-Stream}, the full model with $\mathcal{L}_{\mathrm{stream}}$
($\lambda_{\mathrm{stream}}=0.05$).

\paragraph{Evaluation.}
We evaluate recommendation quality using Recall@$k$ (R@$k$) and Normalized
Discounted Cumulative Gain@$k$ (N@$k$). The main comparisons report R@10, R@50,
N@10, and R@200; additional reported cutoffs are provided in
Appendix~\ref{app:standard_results}. The \textbf{standard} protocol evaluates
positions $t \in [1200,2000)$, yielding 800 prediction steps per user, and is
used unless otherwise specified. The \textbf{extended} protocol evaluates
$t \in [1200,3000)$. A long-history analysis (Section~\ref{sec:analysis})
further evaluates the trained RecMem-Stream model through 10\,000 interactions
on a fixed cohort of 6\,442 users; it compares positional intervals within that
cohort rather than absolute scores across protocols.

\paragraph{Implementation Details.}
Models use embedding dimension $d=256$ and eight attention heads. RecMem uses
the default SplitTop-8/8/8 backbone (Section~\ref{sec:method}), $M=16$ memory
slots, and $L_s=L_r=200$, resulting in 216 decoder-side tokens. PHSTU uses
segment length and recent-window length 200. Unique trainable parameter counts
are reported with the capacity controls (Appendix~\ref{app:capacity_controls}).
Training uses AdamW~\citep{loshchilov2017decoupled} with learning rate
$1.5\times10^{-3}$, weight decay $0.1$, 10\% linear warmup followed by cosine
decay, and gradient clipping at $1.0$. The rollout curriculum increases
$K_{\mathrm{roll}}$ from 1 to 5 while adjusting burn-in within the fixed segment
budget, as described in Section~\ref{sec:method}. Full configurations are
provided in Appendix~\ref{app:hyperparams}.

\subsection{Main Results}
\label{sec:main_results}

\begin{table}[htbp]
\caption{Recommendation quality on MerRec under the standard protocol,
$t\in[1200,2000)$. Scores are percentages, displayed to two decimal places.
Decoder context counts tokens processed per prediction. A dash denotes an
unreported metric.}
\label{tab:main_results}
\centering
\small
\begin{tabular}{lccccc}
\toprule
\textbf{Method} & \textbf{Dec.\ context} & \textbf{R@10} & \textbf{R@50} & \textbf{N@10} & \textbf{R@200} \\
\midrule
HSTU-1200       & 1200     & 48.18 & 63.88 & 31.62 & 74.44 \\
HSTU-FullPrefix & $t$      & 48.98 & 65.64 & 32.32 & ---   \\
PHSTU          & 215--227 & 32.74 & 48.58 & 20.45 & 61.10 \\
HSTU-Recent-200 & 200      & 40.93 & 56.59 & 26.16 & 67.74 \\
Ret-HSTU-16    & 216      & 41.03 & 57.54 & 26.53 & 68.84 \\
RecMem-Main    & 216      & 42.27 & 59.95 & 26.66 & 71.96 \\
RecMem-Stream  & 216      & 42.36 & 60.12 & 26.74 & 72.20 \\
\bottomrule
\end{tabular}
\end{table}

Table~\ref{tab:main_results} compares recommendation quality under the standard
protocol. RecMem-Stream uses 216 decoder-side tokens, $5.6\times$ fewer than
HSTU-1200, while retaining 87.9\% of HSTU-1200's R@10 and 97.0\% of its R@200. The smaller
relative gap at larger cutoffs suggests that the compressed memory retains
much of the predictive signal while sacrificing some top-rank precision,
consistent with the fixed-capacity bottleneck interpretation
(Section~\ref{sec:ib}). Against
the growing-prefix HSTU-FullPrefix reference, RecMem-Stream retains 86.5\%,
91.6\%, and 82.8\% of R@10, R@50, and N@10, respectively. These comparisons
quantify the accuracy--context trade-off of maintaining a fixed recurrent state.

Under the standard protocol, RecMem-Stream exceeds PHSTU by 9.622 and 11.096
percentage points in R@10 and R@200, and the separately trained HSTU-Recent-200
control by 1.433 and 4.454 points, respectively.
Ret-HSTU-16 improves on the recent-only control, showing that archived
interactions provide useful predictive information. RecMem-Stream further
exceeds this retrieval baseline by 1.332/2.575/0.210/3.357 points on
R@10/R@50/N@10/R@200. Differences are computed before rounding the displayed
table values.

\subsection{In-depth Analysis}
\label{sec:analysis}

\paragraph{Capacity controls.}
We evaluate two controls with fewer unique backbone blocks than the default
SplitTop-8/8/8. SharedStack-16 reuses a single 16-block backbone for
memory updating and decoding, whereas SplitTop-6/5/5 keeps separate upper
branches with 16 unique blocks and 11 active blocks per path; shared parameters
are counted once. SharedStack-16 achieves 41.568 R@10 and 70.556 R@200;
SplitTop-6/5/5 achieves 42.055 and 71.354, respectively. Both retain substantial
gains over PHSTU under near-matched parameter counts. The default SplitTop-8/8/8
model provides an additional margin, reaching 42.362 R@10 and 72.196 R@200
(Appendix~\ref{app:capacity_controls}).

\paragraph{Predictive value of distant history.}
\label{sec:ablation_memory}
We compare RecMem with the separately trained HSTU-Recent-200 control on matched
prediction anchors. An anchor is \emph{archived-only} when the target item
previously occurred outside the recent 200 interactions but did not occur
inside that window. We stratify these anchors by whether the last occurrence
lies in the newest 20\%, middle 60\%, or oldest 20\% of the visible prefix,
i.e., all interactions preceding the target position.
Both RecMem variants improve over the recent-only control across all three
strata. On the oldest-20\% anchors, Main and Stream improve R@10 by 0.984 and
1.188 percentage points, with 95\% user-clustered bootstrap intervals of
$[0.931,1.037]$ and $[1.133,1.241]$, respectively. These results show that
RecMem draws predictive signal from archived exact-item history, including its
oldest part, within the evaluated horizon (Appendix~\ref{app:distant_history}).
An inference-time memory-zeroing ablation is reported in
Appendix~\ref{app:memory_zeroing}.

\paragraph{Long-history evaluation.}
\label{sec:trajectory}
Under the extended protocol $t\in[1200,3000)$, RecMem-Stream achieves 41.58 R@10
and 70.96 R@200 while maintaining 216 decoder-side tokens. We further evaluate
the same trained checkpoint through 10\,000 interactions on a fixed cohort of
6\,442 users without retraining. Across four positional intervals,
R@10 ranges from 40.519 to 40.689 and R@200 from 69.867 to 70.379. Two separately
executed evaluations, through 5k and 10k interactions, produce identical
per-position results on their 3\,800 overlapping positions.
Within this fixed cohort, we observe no recommendation-quality collapse or
empirical sign of capacity saturation, long-term forgetting, or
accumulated-update instability through 10\,000 interactions, with memory fixed
at 16 slots. Detailed results are provided in
Appendices~\ref{app:extended_results} and~\ref{app:trajectory}.

\paragraph{Ablations and independent runs.}
In the evaluated write-mechanism ablation, only the gated-residual update
maintains slot diversity (cosine similarity 0.891); the three alternatives
produce effective slot collapse (cosine similarity approximately one).
The memory-capacity sweep
favors $M=16$ among the tested settings, and curriculum training improves R@50
by 1.83 percentage points. Stream consistency improves all four reported
metrics in both seeds 42 and 123, with R@10 gains of 0.089 and 0.813 points,
respectively. Detailed ablations and per-seed results are provided in
Appendices~\ref{app:ablations} and~\ref{app:independent_runs}. We separately
report directional-gain and composition-gap diagnostics to characterize update
behavior under the conditional theoretical framework
(Appendix~\ref{app:stability}).

\paragraph{Model computation.}
A controlled model-computation benchmark (Appendix~\ref{app:systems})
separately measures cold state construction, recurrent updates, and warm
prediction. At batch size 64 and history length 2\,000, RecMem retains 8\,KiB of
state per user (PHSTU: 432\,KiB) and has a warm latency of 7.653\,ms, compared
with 9.245\,ms for PHSTU and 106.613\,ms for HSTU-1200.

\vspace{-1ex}
\section{Conclusion}
\label{sec:conclusion}

We presented RecMem, a recurrent memory compression framework for long-sequence
generative recommendation designed around space invariance, semantic fidelity,
and lifelong evolvability. Its bifurcated encoder--decoder and slot-wise
forget-gated updates maintain a fixed recurrent state and decoder context as
history grows. Under contraction and approximate semigroup consistency, our
analysis bounds the memory-state discrepancy between sequential and packed
processing independently of history length. On MerRec, comparisons with
recent-only and retrieval baselines, together with recency-stratified analyses, demonstrate the
predictive value of archived history, while capacity controls show that the gains
over PHSTU persist with near-matched parameter counts. Evaluating one trained
RecMem-Stream checkpoint on a fixed cohort without retraining shows stable
aggregate recommendation quality through 10,000 interactions. These results
support recurrent memory compression as a practical approach to retaining useful
historical information within a fixed state and context budget.

\begin{ack}
This work was supported by National Natural Science Foundation of China 62502310 and by Alibaba Group through Alibaba Innovative Research Program.
\end{ack}

\bibliographystyle{plainnat}
\bibliography{refs}

\newpage
\appendix
\section{Proof of Theorem~\ref{thm:rollout}}
\label{app:proof_rollout}

We prove Theorem~\ref{thm:rollout} under Assumptions~\ref{ass:contraction} and~\ref{ass:semigroup}. Throughout this proof, state discrepancies are measured using the Frobenius norm. Let $U$ denote the memory update
operator and consider a sequence of $K$ segments $S_1, \ldots, S_K$. Define
the sequential rollout as $M_0^{\mathrm{seq}} = \mathbf{0}$ and
$M_k^{\mathrm{seq}} = U(M_{k-1}^{\mathrm{seq}}, S_k)$ for $k = 1, \ldots, K$.
Let $P_k = S_1 \oplus \cdots \oplus S_k$ denote the packed prefix
($P_0=\emptyset$), and define the packed-prefix states
$M_k^{\mathrm{pack}} = U(\mathbf{0}, P_k)$ with
$M_0^{\mathrm{pack}} = \mathbf{0}$. Thus
$M_K^{\mathrm{pack}}$ is the single-pass packed computation over the full
prefix.

We proceed by induction on the number of processed segments. The base case is
$e_0 = \|M_0^{\mathrm{seq}} - M_0^{\mathrm{pack}}\| = 0$. For the inductive
step, assume $e_k \le \varepsilon \sum_{j=0}^{k-1} \rho^j$. Consider the
error at step $k+1$:
\begin{align}
e_{k+1}
&= \| M_{k+1}^{\mathrm{seq}} - M_{k+1}^{\mathrm{pack}} \| \notag \\
&= \| U(M_k^{\mathrm{seq}}, S_{k+1}) - U(M_k^{\mathrm{pack}}, S_{k+1})
     + U(M_k^{\mathrm{pack}}, S_{k+1}) - M_{k+1}^{\mathrm{pack}} \| \notag \\
&\le \| U(M_k^{\mathrm{seq}}, S_{k+1}) - U(M_k^{\mathrm{pack}}, S_{k+1}) \|
     + \| U(M_k^{\mathrm{pack}}, S_{k+1}) - M_{k+1}^{\mathrm{pack}} \|.
\label{eq:triangle}
\end{align}

By Assumption~\ref{ass:contraction} (contraction) and the mean-value
inequality, the first term satisfies
\begin{equation}
\| U(M_k^{\mathrm{seq}}, S_{k+1}) - U(M_k^{\mathrm{pack}}, S_{k+1}) \|
\le \rho \cdot e_k.
\label{eq:contraction_step}
\end{equation}

The second term measures the discrepancy between processing segment $S_{k+1}$
after the packed prefix state $M_k^{\mathrm{pack}} = U(\mathbf{0}, P_k)$
versus packing $S_{k+1}$ into the prefix itself:
\begin{equation}
\| U(M_k^{\mathrm{pack}}, S_{k+1}) - M_{k+1}^{\mathrm{pack}} \|
= \| U(U(\mathbf{0}, P_k), S_{k+1})
     - U(\mathbf{0}, P_k \oplus S_{k+1}) \|.
\label{eq:packed_prefix_gap}
\end{equation}
For $k=0$, this term is zero. For $k \ge 1$,
Assumption~\ref{ass:semigroup} (approximate semigroup consistency) applies to
the packed prefix block $P_k$ and the next segment $S_{k+1}$, giving
\begin{equation}
\| U(U(\mathbf{0}, P_k), S_{k+1})
     - U(\mathbf{0}, P_k \oplus S_{k+1}) \|
\le \varepsilon.
\label{eq:semigroup_step}
\end{equation}

Combining \eqref{eq:triangle}, \eqref{eq:contraction_step}, and
\eqref{eq:semigroup_step}:
\[
e_{k+1} \le \rho \cdot e_k + \varepsilon.
\]

Unrolling the recursion from $e_0 = 0$:
\[
e_K \le \varepsilon \sum_{j=0}^{K-1} \rho^j
     = \varepsilon \cdot \frac{1 - \rho^K}{1 - \rho}
     \le \frac{\varepsilon}{1 - \rho},
\]
where the last inequality uses $\rho^K \ge 0$. \qed

\section{Hyperparameters}
\label{app:hyperparams}

Table~\ref{tab:hyperparams} summarizes the RecMem training configuration.

\begin{table}[htbp]
\caption{Full hyperparameter configuration for RecMem on MerRec.}
\label{tab:hyperparams}
\centering
\begin{tabular}{ll}
\toprule
\textbf{Parameter} & \textbf{Value} \\
\midrule
Embedding dimension $d$ & 256 \\
Unique HSTU backbone blocks & 24 \\
Active blocks per path & 16 \\
Shared lower layers & 8 \\
Encoder upper layers & 8 \\
Decoder upper layers & 8 \\
Attention heads & 8 \\
Memory slots $M$ & 16 \\
Segment length $L_s$ & 200 \\
Recent window length $L_r$ & 200 \\
Memory norm bound $c$ & 10.0 \\
\midrule
Optimizer & AdamW \\
Learning rate & $1.5 \times 10^{-3}$ \\
Weight decay & 0.1 \\
LR schedule & Linear warmup + cosine decay \\
Warmup ratio & 10\% \\
Max gradient norm & 1.0 \\
Batch size (RecMem / HSTU-1200) & 64 / 32 \\
\midrule
Burn-in segments $K_\text{burn}$ & $K-K_\text{roll}$ (curriculum) \\
Rollout segments $K_\text{roll}$ & 1 $\to$ 5 (curriculum) \\
Curriculum switch steps & 6990, 16310, 27960, 39610 \\
\midrule
$\lambda_\text{stream}$ (stream consistency) & 0.05 \\
\midrule
Training history length & 1200 \\
Total sequence length & 2000 \\
Default random seed & 42 \\
Independent Main/Stream run seed & 123 \\
\bottomrule
\end{tabular}
\end{table}

\section{Additional Standard Evaluation Results}
\label{app:standard_results}

Table~\ref{tab:standard_full} reports all evaluated cutoffs for HSTU-1200, PHSTU,
and the two RecMem variants under the standard protocol;
Table~\ref{tab:main_results} compares all baselines on four metrics.

\begin{table}[htbp]
\caption{All evaluated cutoffs for HSTU-1200, PHSTU, and RecMem under the standard protocol,
$t\in[1200,2000)$. Scores are percentages. Best per metric in \textbf{bold};
best among compressed methods \underline{underlined}.}
\label{tab:standard_full}
\centering
\resizebox{\textwidth}{!}{%
\begin{tabular}{lccccccccccc}
\toprule
\textbf{Method} & \textbf{Dec.\ context} &
\textbf{R@5} & \textbf{R@10} & \textbf{R@50} & \textbf{R@100} & \textbf{R@200} &
\textbf{N@5} & \textbf{N@10} & \textbf{N@50} & \textbf{N@100} & \textbf{N@200} \\
\midrule
HSTU-1200 & 1200
  & \textbf{39.69} & \textbf{48.18} & \textbf{63.88} & \textbf{69.45} & \textbf{74.44}
  & \textbf{28.86} & \textbf{31.62} & \textbf{35.15} & \textbf{36.05} & \textbf{36.75} \\
\midrule
PHSTU & 215--227
  & 25.56 & 32.74 & 48.58 & 54.99 & 61.10
  & 18.13 & 20.45 & 23.99 & 25.03 & 25.88 \\
RecMem-Main & 216
  & 33.41 & 42.27 & 59.95 & 66.30 & 71.96
  & 23.79 & 26.66 & 30.63 & 31.66 & 32.45 \\
RecMem-Stream & 216
  & \underline{33.48} & \underline{42.36} & \underline{60.12} & \underline{66.51} & \underline{72.20}
  & \underline{23.87} & \underline{26.74} & \underline{30.73} & \underline{31.77} & \underline{32.56} \\
\bottomrule
\end{tabular}%
}
\end{table}

\section{Capacity Controls}
\label{app:capacity_controls}

Table~\ref{tab:capacity_controls} reports unique trainable parameter counts,
counting shared parameters once. SharedStack-16 uses the same 16-block backbone
for memory updating and decoding. SplitTop-6/5/5 has six shared, five
encoder-specific, and five decoder-specific blocks; its 16 unique blocks give
11 active blocks per path. Both controls retain substantial gains over PHSTU
under near-matched parameter counts. The default SplitTop-8/8/8 architecture
has 24 unique blocks and 16 active blocks per path.

\begin{table}[htbp]
\caption{Capacity controls under the standard protocol. Scores are percentages;
parameter counts are in millions. Shared parameters are counted once.}
\label{tab:capacity_controls}
\centering
\footnotesize
\setlength{\tabcolsep}{4pt}
\begin{tabular}{lrrrrrrr}
\toprule
\textbf{Model} & \textbf{Params.\ (M)} & \textbf{Unique blocks} & \textbf{Active/path} & \textbf{R@10} & \textbf{R@50} & \textbf{N@10} & \textbf{R@200} \\
\midrule
PHSTU & 327.217 & 16 & 16 & 32.740 & 48.580 & 20.450 & 61.100 \\
SharedStack-16 & 327.283 & 16 & 16 & 41.568 & 58.611 & 26.505 & 70.556 \\
SplitTop-6/5/5 & 327.342 & 16 & 11 & 42.055 & 59.471 & 26.326 & 71.354 \\
SplitTop-8/8/8 (default) & 329.906 & 24 & 16 & 42.362 & 60.119 & 26.743 & 72.196 \\
\bottomrule
\end{tabular}
\end{table}

\section{Retrieval Baseline}
\label{app:retrieval}

Ret-HSTU-16 is inspired by retrieval-augmented sequential recommendation
\citep{wu2023ruel}. It retrieves from the current user's own archived history.
At each prediction anchor, a frozen HSTU-Recent-200 encodes the preceding 200
interactions; its normalized final causal hidden state serves as the query.
Exact FP32 cosine search ranks archived item occurrences using the frozen
retriever's item embeddings. The top 16 occurrences are selected without
deduplicating item IDs, with exact-score ties favoring newer positions, then
restored to chronological order and prepended to the recent window. The
predictor therefore receives 216 tokens. The archive is restricted to positions
strictly before the recent window, and neither the query nor the archive has
access to the target interaction or future interactions. Table~\ref{tab:main_results}
compares this baseline with the separately trained recent-only control and
RecMem under the standard protocol.

\section{Predictive Value of Distant History}
\label{app:distant_history}

We compare predictions at matched (user, target-position) anchors without
changing training or inference. For each anchor, we locate the target item's
last prior occurrence. An archived-only anchor has a prior occurrence outside
the recent 200 interactions and none inside that window. We stratify these
anchors by whether that occurrence falls in the newest 20\%, middle 60\%, or
oldest 20\% of the visible prefix, i.e., all non-padding interactions preceding
the target position. Table~\ref{tab:distant_history} reports
Recall@10 differences relative to the separately trained HSTU-Recent-200
control. Both Main and Stream improve across all three recency strata, including
the oldest archived exact-item history within the evaluated horizon.

\begin{table}[htbp]
\caption{Recall@10 differences relative to HSTU-Recent-200 on matched anchors,
in percentage points. Brackets give 95\% confidence intervals from 2\,000
user-clustered bootstrap resamples. Archived-only aggregates the three
archived-history strata; no-recurrence anchors have no prior target-item
occurrence in the visible prefix.}
\label{tab:distant_history}
\centering
\small
\begin{tabular}{lrrr}
\toprule
\textbf{Anchors} & \textbf{Count} & \textbf{Main $\Delta$R@10 [CI]} & \textbf{Stream $\Delta$R@10 [CI]} \\
\midrule
Archived-newest & 2\,898\,654 & +4.413 [4.368, 4.460] & +4.576 [4.532, 4.623] \\
Archived-middle & 7\,593\,974 & +2.169 [2.145, 2.194] & +2.353 [2.327, 2.378] \\
Archived-oldest & 1\,044\,050 & +0.984 [0.931, 1.037] & +1.188 [1.133, 1.241] \\
\midrule
Archived-only & 11\,536\,678 & +2.626 [2.605, 2.647] & +2.806 [2.785, 2.827] \\
No-recurrence & 20\,786\,602 & +0.318 [0.309, 0.327] & +0.433 [0.423, 0.443] \\
\bottomrule
\end{tabular}
\end{table}

\subsection{Inference-Time Memory Zeroing}
\label{app:memory_zeroing}

To quantify the contribution of the recurrent memory, we evaluate a
\emph{no-memory} variant that zeros out all $M = 16$ memory slots at each
segment boundary during inference, forcing the model to rely solely on the
recent window. We use a $d = 64$ model for this ablation to isolate the
memory contribution at lower capacity. Table~\ref{tab:memory_importance} reports results
at two inference horizons. This inference-time intervention measures the trained
model's dependence on its memory state and is distinct from the separately
trained recent-only control.
\begin{table}[htbp]
\caption{Memory importance ablation. The no-memory variant zeros out all memory slots at inference, retaining only the recent window.}
\label{tab:memory_importance}
\centering
\begin{tabular}{llcccc}
\toprule
\textbf{Condition} & \textbf{Infer $L$} &
\textbf{R@5} & \textbf{R@10} & \textbf{R@50} & \textbf{N@10} \\
\midrule
RecMem (full) & 1\,000
  & \textbf{33.32} & \textbf{41.48} & \textbf{57.88} & \textbf{26.72} \\
RecMem (no memory) & 1\,000
  & 22.03 & 28.62 & 44.84 & 17.78 \\
\midrule
RecMem (full) & 2\,000
  & \textbf{33.23} & \textbf{41.35} & \textbf{57.69} & \textbf{26.63} \\
RecMem (no memory) & 2\,000
  & 15.17 & 20.23 & 34.14 & 12.10 \\
\bottomrule
\end{tabular}
\end{table}

\section{Extended Evaluation Results}
\label{app:extended_results}

Table~\ref{tab:extended_results} reports aggregate results for
$t\in[1200,3000)$. RecMem-Main and RecMem-Stream achieve R@200 of 70.67 and
70.96, respectively, compared with 64.02 for PHSTU and 73.87 for HSTU-1200.
RecMem-Stream therefore retains approximately 96\% of HSTU-1200's R@200 while
using a fixed 216-token decoder context. The 10k analysis uses a separate
fixed cohort, so its absolute scores are interpreted through comparisons among
intervals within that cohort.

\begin{table}[htbp]
\caption{Extended evaluation on MerRec (test positions $t \in [1200, 3000)$,
1\,800 steps, up to 2.5$\times$ training horizon).
Best per metric in \textbf{bold}; best among compressed methods \underline{underlined}.}
\label{tab:extended_results}
\centering
\resizebox{\textwidth}{!}{%
\begin{tabular}{lccccccccccc}
\toprule
\textbf{Method} & \textbf{Dec.\ context} &
\textbf{R@5} & \textbf{R@10} & \textbf{R@50} & \textbf{R@100} & \textbf{R@200} &
\textbf{N@5} & \textbf{N@10} & \textbf{N@50} & \textbf{N@100} & \textbf{N@200} \\
\midrule
HSTU-1200 & 1200
  & \textbf{39.80} & \textbf{48.12} & \textbf{63.43} & \textbf{68.92} & \textbf{73.87}
  & \textbf{29.00} & \textbf{31.70} & \textbf{35.15} & \textbf{36.04} & \textbf{36.73} \\
\midrule
PHSTU & 215--227
  & 30.29 & 37.64 & 52.62 & 58.48 & 64.02
  & 21.79 & 24.17 & 27.52 & 28.48 & 29.25 \\
RecMem-Main & 216
  & 32.81 & 41.45 & 58.71 & 65.01 & 70.67
  & 23.41 & 26.21 & 30.08 & 31.10 & 31.89 \\
RecMem-Stream & 216
  & \underline{32.92} & \underline{41.58} & \underline{58.94} & \underline{65.27} & \underline{70.96}
  & \underline{23.50} & \underline{26.31} & \underline{30.20} & \underline{31.23} & \underline{32.03} \\
\bottomrule
\end{tabular}%
}
\end{table}

\section{Long-History Evaluation through 10k Interactions}
\label{app:trajectory}

Table~\ref{tab:long_history} reports the same RecMem-Stream checkpoint on a
fixed cohort of 6\,442 users, without retraining. Its 16-slot memory and
216-token decoder context remain unchanged. Separately executed 5k and 10k
evaluations produce identical per-position results on all 3\,800 overlapping
positions.

\begin{table}[htbp]
\caption{Long-history evaluation of the same RecMem-Stream checkpoint on a
fixed cohort of 6\,442 users. Scores are percentages aggregated within each
positional interval. Memory size remains fixed at 16 slots, and no retraining
is performed.}
\label{tab:long_history}
\centering
\begin{tabular}{lrrrr}
\toprule
\textbf{Positions $t$} & \textbf{R@10} & \textbf{R@50} & \textbf{N@10} & \textbf{R@200} \\
\midrule
1200--1999 & 40.519 & 57.736 & 25.632 & 69.867 \\
2000--4999 & 40.644 & 57.911 & 25.744 & 70.032 \\
5000--7499 & 40.684 & 57.999 & 25.741 & 70.154 \\
7500--9999 & 40.689 & 58.137 & 25.726 & 70.379 \\
\bottomrule
\end{tabular}
\end{table}

\begin{figure}[htbp]
\centering
\includegraphics[width=0.9\linewidth]{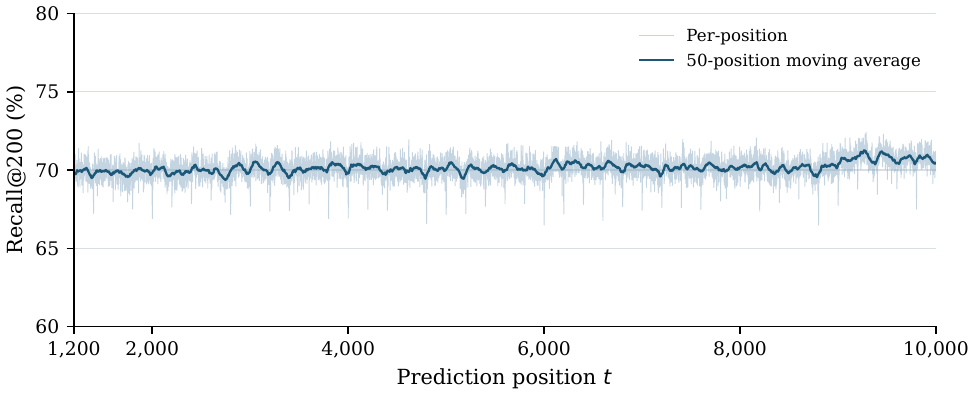}
\caption{Recall@200 of RecMem-Stream through 10\,000 interactions on a fixed
cohort of 6\,442 users, using the same trained checkpoint and 16 memory slots
without retraining. The light curve shows per-position values; the dark curve
is a centered 50-position moving average, using only available positions at
the boundaries.}
\label{fig:trajectory}
\end{figure}

Figure~\ref{fig:trajectory} plots RecMem-Stream's Recall@200 over positions
$t\in[1200,10000)$ for this fixed cohort. The trajectory complements the
interval aggregates by showing how recommendation quality varies as additional
historical segments are processed. It measures predictive behavior over the
observed horizon; the state-space stability theorem remains conditional on
its stated assumptions.

\section{Independent Main/Stream Runs}
\label{app:independent_runs}

Table~\ref{tab:independent_runs} reports Main and Stream separately for seeds
42 and 123 under the standard protocol. Stream improves all four metrics in
both paired comparisons, although the gain magnitude varies across seeds.
These results show a positive Stream-over-Main direction across both evaluated
seeds.

\begin{table}[htbp]
\caption{Individual Main/Stream runs under the standard protocol. Scores are
percentages.}
\label{tab:independent_runs}
\centering
\begin{tabular}{llrrrr}
\toprule
\textbf{Seed} & \textbf{Variant} & \textbf{R@10} & \textbf{R@50} & \textbf{N@10} & \textbf{R@200} \\
\midrule
42 & Main   & 42.273 & 59.952 & 26.659 & 71.962 \\
42 & Stream & 42.362 & 60.119 & 26.743 & 72.196 \\
\midrule
123 & Main   & 42.457 & 60.156 & 26.791 & 72.152 \\
123 & Stream & 43.270 & 60.848 & 27.440 & 72.729 \\
\bottomrule
\end{tabular}
\end{table}

\section{Empirical Update Diagnostics}
\label{app:stability}
\label{app:stability_verification}

We separately examine sensitivity to memory perturbations and discrepancies
caused by regrouping segments. These measurements characterize evaluated
rollouts, while Theorem~\ref{thm:rollout} remains conditional on contraction and
approximate semigroup consistency.

Our first probe estimates sampled directional gain using randomized finite
differences with perturbation size $\delta=10^{-3}$, five update positions in
each $K=5$ rollout, and 20 validation batches. Main and Stream yield
$\widehat{\rho}_{\mathrm{dir}}=0.630$ and $0.638$, respectively, indicating
attenuation along the sampled directions. This directional estimate is
distinct from a worst-direction Jacobian spectral norm and does not certify
uniform contraction.

The expanded composition diagnostic covers Main and Stream, seeds 42 and 123,
and 160 users per evaluated cell. We normalize the Frobenius gap between two
sequential updates and one packed update by $\sqrt{Md}$. At $K=5$, median gaps
range from 0.079 to 0.103 and p95 values from 0.111 to 0.147. Across
$K\in\{5,9,14\}$, median gaps remain between 0.079 and 0.105. These are empirical
local composition gaps, rather than uniform bounds on the theorem's
$\varepsilon$.

On paired users, Spearman point estimates between Main-to-Stream reductions in
output KL and state discrepancy range from 0.02 to 0.45. The association varies
across checkpoints and horizons, and the average directions of the two changes
are not jointly consistent across seeds. We therefore interpret output KL as
a training signal for prediction consistency rather than a general guarantee
on memory-state distance. Table~\ref{tab:stability} summarizes these diagnostics.

\begin{table}[htbp]
\caption{Empirical update diagnostics. Composition gaps are normalized by
$\sqrt{Md}$. Ranges span evaluated variant/seed cells or checkpoint/horizon
cells as indicated; they are not confidence intervals or uniform bounds.}
\label{tab:stability}
\centering
\small
\begin{tabular}{p{0.40\linewidth}p{0.33\linewidth}r}
\toprule
\textbf{Diagnostic} & \textbf{Scope} & \textbf{Value} \\
\midrule
Sampled directional gain & Original probe: Main / Stream & 0.630 / 0.638 \\
Composition-gap median & $K=5$, variants and seeds & 0.079--0.103 \\
Composition-gap p95 & $K=5$, variants and seeds & 0.111--0.147 \\
Composition-gap median & $K\in\{5,9,14\}$, variants and seeds & 0.079--0.105 \\
Spearman correlation & Evaluated checkpoints and horizons & 0.02--0.45 \\
\bottomrule
\end{tabular}
\end{table}

\section{Model-Computation Benchmark}
\label{app:systems}

We benchmark model computation using randomly initialized architectures and
synthetic item-ID inputs on a single GPU in bfloat16, with three warm-up runs
and 20 measured repetitions. Reported latencies are medians measured with GPU
synchronization. Cold history-state construction, a subsequent 200-item update,
and warm prediction are measured separately. Warm full prediction includes
decoder computation, full-catalog candidate scoring, and top-200 selection with
history state already prepared. Measurements exclude data loading, networking,
remote state storage, request scheduling, and other service overhead.

At batch size 64 and available history length 2\,000, RecMem achieves a warm
latency of 7.653\,ms and throughput of 8\,363 users/s, with 1.585\,GB peak
allocated GPU memory (Table~\ref{tab:systems_warm}). Its retained recurrent
state occupies 8\,KiB per user. These measurements describe model computation
under the specified benchmark rather than end-to-end production serving.

\begin{table}[htbp]
\caption{Warm full prediction at batch size 64 and available history length
2\,000. Throughput is batch size divided by median latency. Peak GPU memory is
allocated memory in decimal GB; retained user state is in KiB ($2^{10}$ bytes).
HSTU-1200 uses its latest 1\,200 interactions and has no retained recurrent
user state.}
\label{tab:systems_warm}
\centering
\small
\begin{tabular}{lrrrr}
\toprule
\textbf{Model} & \textbf{Latency (ms)} & \textbf{Users/s} & \textbf{Peak GPU (GB)} & \textbf{State/user (KiB)} \\
\midrule
RecMem & 7.653 & 8363 & 1.585 & 8 \\
PHSTU & 9.245 & 6923 & 1.586 & 432 \\
HSTU-1200 & 106.613 & 600 & 6.285 & None \\
\bottomrule
\end{tabular}
\end{table}

Table~\ref{tab:systems_phases} separates state construction, update, and warm
prediction at batch size 1. Cold construction excludes prediction. Update
timing starts from the same pre-update state for each repetition; one update
is then committed outside the timed region before measuring warm prediction.
Reported state sizes are after that update and exclude model parameters,
item-catalog embeddings, and the raw recent window.

\begin{table}[htbp]
\caption{Separate model-computation phases at batch size 1. Cold denotes history-state
construction; update processes one 200-item segment; warm includes decoder,
full-catalog scoring, and top-200. History $H$ is the post-update horizon;
updates advance from $H-200$ to $H$. Latencies are in ms, and retained state
sizes are in KiB per user after the update.}
\label{tab:systems_phases}
\centering
\begin{tabular}{llrrrr}
\toprule
\textbf{Model} & \textbf{History $H$} & \textbf{Cold (ms)} & \textbf{Update (ms)} & \textbf{Warm (ms)} & \textbf{State (KiB)} \\
\midrule
RecMem & 1200 & 23.1 & 4.57 & 3.82 & 8 \\
RecMem & 2000 & 41.4 & 4.61 & 3.81 & 8 \\
PHSTU & 1200 & 19.8 & 4.11 & 3.99 & 240 \\
PHSTU & 2000 & 35.8 & 4.09 & 4.03 & 432 \\
\bottomrule
\end{tabular}
\end{table}

\section{Ablation Studies}
\label{app:ablations}

\subsection{Memory Write Mechanism}
\label{app:ablation_write}

\begin{table}[htbp]
\caption{Comparison of memory write mechanisms. Slot cosine similarity (Slot Cos) measures pairwise similarity among memory slots; values near 1 indicate effective collapse.}
\label{tab:write_mechanism}
\centering
\begin{tabular}{lcccc}
\toprule
\textbf{Write Mechanism} &
\textbf{R@10} & \textbf{R@50} & \textbf{Slot Cos} & \textbf{Collapse?} \\
\midrule
Gated-residual (ours)
  & \textbf{40.92} & \textbf{59.10} & 0.891 & No \\
Attention-write
  & 40.33 & 58.23 & $\approx$1 & Yes \\
Attention-write ($\tau{=}0.5$)
  & 40.27 & 58.10 & $\approx$1 & Yes \\
Erase-write~\citep{graves2014neural}
  & 40.38 & 58.45 & $\approx$1 & Yes \\
\bottomrule
\end{tabular}
\end{table}

Table~\ref{tab:write_mechanism} compares four memory write mechanisms.
All three alternatives---attention-based writing, temperature-scaled
attention writing, and Neural Turing Machine-style erase-write~\citep{graves2014neural}---produce
effective slot collapse: the pairwise cosine similarity among all 16 slots
approaches 1, indicating nearly identical slot content that effectively degenerates to a single repeated vector.
Only the gated-residual update maintains slot diversity (cosine similarity
0.891). While the accuracy gap is moderate (0.54--0.65 percentage points in Recall@10),
slot collapse sharply reduces the effective memory capacity: $M = 16$
slots carry nearly identical content, defeating the purpose of a
multi-slot memory. The gated-residual design avoids this by blending new
information with the existing state through a learned forget gate, preserving
the per-slot specialization needed for semantic fidelity.

\subsection{Memory Capacity and Curriculum Training}
\label{app:ablation_slots}
\label{app:ablation_curriculum}

\begin{table}[htbp]
\caption{(a)~Effect of memory slot count $M$. (b)~Effect of curriculum training.}
\label{tab:slots_ablation}
\label{tab:curriculum}
\centering
\small
\begin{subtable}[t]{0.46\linewidth}
\centering\footnotesize
\begin{tabular}{ccccc}
\toprule
$M$ & \textbf{R@10} & \textbf{R@50} & \textbf{N@10} & \textbf{N@50} \\
\midrule
8  & 33.65 & 45.22 & 22.27 & 24.84 \\
16 & \textbf{35.51} & \textbf{46.38} & \textbf{23.41} & \textbf{25.83} \\
32 & 35.18 & 46.14 & 23.13 & 25.58 \\
\bottomrule
\end{tabular}
\caption{Memory capacity}
\end{subtable}
\hfill
\begin{subtable}[t]{0.52\linewidth}
\centering\footnotesize
\begin{tabular}{lcccc}
\toprule
 & \textbf{R@5} & \textbf{R@10} & \textbf{R@50} & \textbf{N@10} \\
\midrule
Curriculum    & \textbf{34.85} & \textbf{43.64} & \textbf{61.29} & \textbf{27.89} \\
No curriculum & 34.43 & 43.05 & 59.46 & 27.40 \\
\bottomrule
\end{tabular}
\caption{Curriculum training}
\end{subtable}
\end{table}

Table~\ref{tab:slots_ablation}a reports the memory-capacity sweep with
$L_s=L_r=50$. Among the tested settings, $M=16$ achieves the best accuracy on
all four metrics while using half the slots of $M=32$; its R@10 exceeds $M=8$
by 1.86 percentage points. Table~\ref{tab:slots_ablation}b reports a 1.83-point
R@50 improvement from curriculum training~\citep{bengio2009curriculum} in the
evaluated ablation setting. These results characterize the empirical effects
of memory capacity and rollout training.

\section{Training Algorithm}
\label{app:algorithm}
\begin{algorithm}[H]
\caption{RecMem Training (one batch)}
\label{alg:training}
\small
\begin{algorithmic}[1]
\REQUIRE Segments $\{S_1, \ldots, S_K\}$, recent suffix $X_{\mathrm{rec}}$, positives $Y^{+}$, negatives $Y^{-}$
\REQUIRE Burn-in count $K_\text{burn}$, rollout count $K_\text{roll}$
\STATE $M_0 \leftarrow \mathbf{0}$ \COMMENT{Initialize memory}
\STATE \textbf{// Burn-in: no gradients}
\FOR{$t = 1$ to $K_\text{burn}$}
    \STATE $M_t \leftarrow U(M_{t-1}, S_t)$ without gradients
\ENDFOR
\STATE $M_{K_\text{burn}} \leftarrow \text{detach}(M_{K_\text{burn}})$
\STATE \textbf{// Rollout: with gradients}
\FOR{$t = K_\text{burn}+1$ to $K$}
    \STATE $M_t \leftarrow U(M_{t-1}, S_t)$ with gradients
\ENDFOR
\STATE \textbf{// Decode and compute losses}
\STATE $H \leftarrow \text{Decoder}(M_K, X_{\mathrm{rec}})$
\STATE $\mathcal{L}_\text{main} \leftarrow \text{NCE}(H, Y^{+}, Y^{-})$
\STATE \textbf{// Stream consistency: pack last two segments}
\STATE $M_\text{pack} \leftarrow U(M_{K-2},\; S_{K-1} \oplus S_K)$
\STATE $H_\text{pack} \leftarrow \text{detach}(\text{Decoder}(M_\text{pack}, X_{\mathrm{rec}}))$
\STATE $\mathcal{L}_\text{stream} \leftarrow \mathrm{KL}(p(H_\text{pack}) \,\|\, p(H))$
\STATE $\mathcal{L} \leftarrow \mathcal{L}_\text{main} + \lambda_\text{stream} \mathcal{L}_\text{stream}$
\STATE Update parameters with AdamW
\end{algorithmic}
\end{algorithm}

\section{Limitations}
\label{app:limitations}

Our evaluation focuses on ID-based recommendation on MerRec with HSTU-family
backbones. Fixed state size and decoder context follow from the architecture,
while predictive retention is supported over the evaluated histories and the
rollout discrepancy bound depends on its stated assumptions. The 10k evaluation
uses one trained Stream checkpoint on a fixed cohort, and the Main/Stream
comparison covers two paired seeds. The systems results measure model
computation rather than production serving. Our Rec2PM-O reproduction did not
yield a reliable quantitative comparison, so we make no empirical superiority
claim against it. Future work includes additional datasets and backbones,
adaptive segment or slot allocation, and richer textual or multimodal inputs.

\clearpage


\end{document}